\documentclass[letterpaper, 10 pt, conference]{ieeeconf}
\IEEEoverridecommandlockouts
\usepackage{amsmath,amssymb}
\usepackage{graphicx}
\usepackage{booktabs}
\usepackage{times}
\usepackage{xurl}

\title{\bf Shaft-Configuration-Adaptive Catheter Tip Position Estimation \\ via Motor-History Conditioned Residual Learning}

\author{
	Peihan Zhang$^{1,2}$,
	Michael C. Yip$^{2}$,
	Ankur Kapoor$^{1}$,
	and Young-Ho Kim$^{1}$%
	\thanks{$^{1}$Digital Technology \& Innovation, Siemens Healthineers,
		Princeton, NJ, USA.}
	\thanks{$^{2}$Department of Electrical and Computer Engineering,
		University of California San Diego, United States.}
	\thanks{\tt\small young-ho.kim@siemens-healthineers.com}
}
\begin{document}
\maketitle

\begin{abstract}
Tendon-driven continuum manipulators are widely used in medical applications, where accurate tip-position estimation is essential for precise navigation and instrument positioning.
However, patient anatomy and procedural setup impose task-dependent unknown shaft configurations, while friction, slack, and compliance introduce hysteresis, making tip estimation challenging.
This paper presents a motor-history-conditioned gated recurrent unit (GRU) residual estimator for three-dimensional catheter tip estimation without direct shaft-configuration sensing.
First, an initial multidirectional sweep strategy is applied to calibrate a geometric catheter model backbone, and encode the motor-angle and drive-torque response into a shaft-configuration context vector.
During subsequent motion, the context conditions a GRU that predicts a task-space residual correcting this backbone, relying on motor measurements alone.
The context remains fixed for the current shaft configuration, while the recurrent state captures the evolving actuation history.
Across four disposable intra-cardiac echocardiography catheters and 16 bent shaft configurations, the method achieves 3.3\,mm open-loop tip RMSE, a 59\% reduction relative to the constant-curvature baseline.
\end{abstract}

\section{Introduction}

\begin{figure}[t]\centering
\includegraphics[width=\columnwidth]{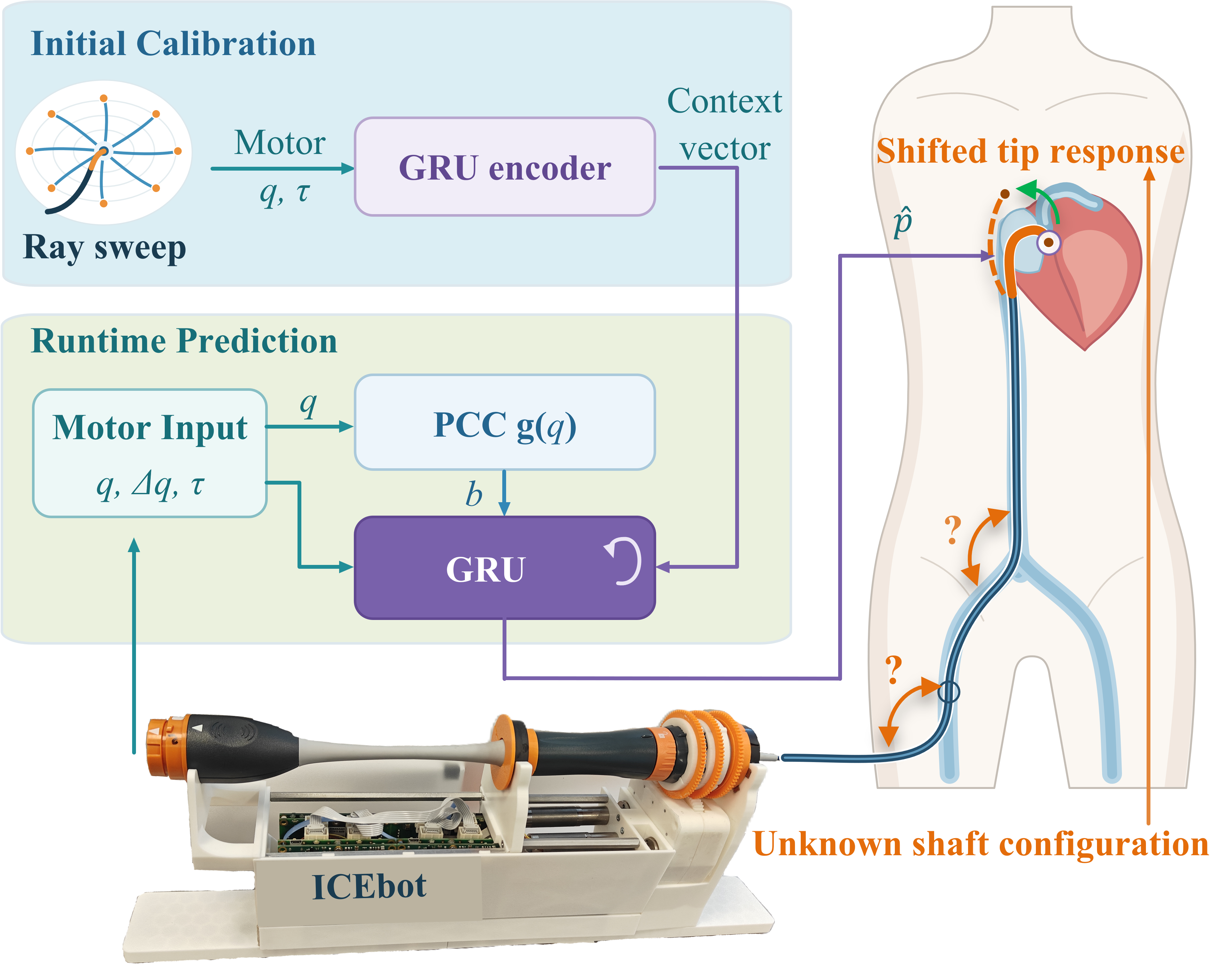}
\caption{Overview of the system workflow.
ICEbot, an intra-cardiac echocardiography (ICE) catheter robot, encounters task-dependent unknown shaft configurations during setup.
An initial multidirectional tip sweep provides motor history for shaft-configuration context encoding.
The GRU residual corrects the geometric catheter model's tip prediction during subsequent estimation from motor measurements alone.}
\label{fig:arch}
\end{figure}

\newcommand{\introductiontext}{%
Tendon-driven continuum manipulators are widely used in minimally invasive medical procedures to navigate constrained anatomical pathways through flexible distal motion and proximal actuation.
Intra-cardiac echocardiography (ICE) catheters are a representative example: tendon-driven tip bending positions a distal ultrasound array for cardiac imaging.
Accurate tip-position estimation therefore supports both catheter navigation and
consistent positioning of the imaging probe.

In clinical scenarios, the shaft configuration is task dependent: patient anatomy, the vascular access route, and procedural setup impose bending directions and
magnitudes that are unknown at run time.
Such configurations alter tendon routing and transmission behavior, so that the same motor input produces a configuration-dependent distal response.
What's more, hysteresis arising from friction, slack, and compliance compounds this difficulty: even for a fixed shaft configuration, the tip position depends on the preceding actuation.

Several strategies have been proposed to address these effects.
One is to measure the shaft configuration directly, through integrated shape sensors such as optical fiber~\cite{gao2025route} or visual observation~\cite{shi2017shape}.
These measurements are informative, but they require additional sensor integration or usable image observations, which are not always accessible on the device.
An alternative is to compensate without shaft sensing, inferring the distal response from proximal actuation alone.
This setting is more demanding, as the estimator must account for both configuration-dependent geometric mismatch and motion-history dependence.
Motor-current-based dead-zone compensation~\cite{lee2021} and configuration-adaptive offset identification~\cite{kim2021} capture important transmission effects without explicit shaft reconstruction.
Learning-based models offer a complementary capability: recurrent architectures such as long short-term memory (LSTM) networks~\cite{wu2021} and GRU networks~\cite{wang2025comparison} carry an internal state across samples, and can therefore represent the dependence of the tip response on preceding actuation without an explicit hysteresis model.
These models, however, are usually fitted to one shaft configuration and must be retrained when it changes.

In this paper, we aim to address both an unknown task-dependent shaft
configuration and actuation-history dependence within a single estimator driven
by motor measurements alone, while maintaining model generalizability across
catheters of the same device family.
We propose a motor-history-conditioned GRU residual estimator coupled with a multidirectional calibration strategy as shown in Fig.~\ref{fig:arch}.
The tip position is estimated as a geometric prediction plus a learned residual:
a geometric catheter model gives the nominal response to the current motor
angles, and a GRU predicts the remaining deviation from the motor history,
comprising motor-angle and drive-torque sequences.
The GRU is additionally conditioned on a shaft-configuration context vector,
encoded from the motor history recorded during the initial sweep.
Different shaft configurations yield different contexts, so the residual adapts
to the configuration in which the catheter is operating.
This adaptation is carried entirely by the context vector and the geometric
parameters, which are identified once per configuration and then held fixed.
The same recurrent network weights therefore apply to any configuration, and to any
catheter of the same device family, since unit- and configuration-specific
behavior is absorbed by the context vector and the geometric parameters rather
than by the learned weights.


\begin{samepage}
The contributions of this paper are:
\begin{itemize}
  \item \textbf{Residual tip estimator from motor measurements alone.}
    A three-dimensional catheter tip estimator that combines a  geometric prediction with a motor-history-conditioned
    recurrent residual, separating the nominal configuration-dependent response
    from its history-dependent deviation. After the initial sweep, estimation requires only motor-angle and drive-torque measurements, without shaft-configuration sensing.
  \item \textbf{Dual-purpose sweep strategy for configuration adaptation.}
    A multidirectional sweep that both calibrates the geometric model and encodes
    a shaft-configuration context vector for the recurrent residual, so that an
    unknown configuration is identified from the same actuation signals later
    used for estimation, and adapting to a new configuration requires only a
    repeated sweep rather than retraining.
  \item \textbf{Validation across catheter units and shaft configurations.}
    Benchtop validation on four disposable catheter units from the same
    device family, to which the identical model architecture and calibration procedure
    are applied, across shaft configurations with unknown bending directions and
    magnitudes, achieving 3.3\,mm open-loop tip-position RMSE and a 59\% reduction relative to baseline.
\end{itemize}
  \end{samepage}
}

\suppressfloats[t]
\introductiontext

\section{Related Work}\label{sec:related-work}

Constant-curvature and physics-based models provide nominal actuation-to-tip mappings~\cite{webster2010,camarillo2008}, whose accuracy can be reduced by shaft-configuration changes and actuation-history dependence.
Various methods address these effects, which we organize into two categories by sensing requirements and prediction objectives: methods that adapt to the shaft configuration, and methods that model actuation-history dependence.

\textbf{Adapting to the shaft configuration.}
One line of work measures the configuration directly.
In~\cite{shi2017shape}, fiber-optic, electromagnetic, and intraoperative imaging approaches to shape reconstruction are reviewed, while~\cite{shentu2024moss} distinguishes embedded from vision-based sensing and reconstructs a continuum robot's centerline from a single RGB image.
Measured shape can also close the control loop: \cite{roesthuis2016fbg} combines fiber Bragg grating measurements with rigid-link modeling for bending control.
Beyond observing robot deformation, measured routing can inform transmission compensation, as in~\cite{gao2025route}, which uses route-sensing optical fiber to compensate tendon elongation and hysteresis under time-varying tendon-sheath configurations, and~\cite{hong2025camera}, which identifies an equivalent-circle representation from endoscopic marker observations.
A recent review~\cite{li2026transmission} organizes such methods by their sensing requirements, distinguishing adaptive feedback, pre-identified feedforward, and route-interpretable feedforward control.
These approaches are effective where such sensing exists, but many clinical catheters offer neither integrated fibers nor a usable view of the shaft.

An alternative infers configuration-dependent transmission effects from proximal signals already available at the actuation unit.
Lee et al.~\cite{lee2021} identify dead-zone and backlash parameters from motor current, and Kim and Mansi~\cite{kim2021} detect configuration-dependent dead-zone shifts from current without reconstructing the shaft, compensating them through a piecewise-linear model.
In~\cite{debuys2023}, hysteresis and re-tension compensation are investigated for a long passive proximal section under changing shaft configurations.

\textbf{Modeling actuation-history dependence.}
Within a fixed configuration, the distal response depends on preceding motion.
Analytical transmission models capture this dependence explicitly: \cite{do2014} models asymmetric tendon-sheath hysteresis, and~\cite{kato2016} incorporates tendon friction into forward kinematics.
Data-driven temporal models learn it instead.
A long short-term memory (LSTM) network predicts catheter bending from proximal pressure in~\cite{wu2021}; \cite{kim2022preisach} combines a Preisach model with a recurrent network for configuration-specific tendon-sheath hysteresis; \cite{wang2024hysteretic} compares feedforward, history-buffer, and LSTM models; \cite{wang2025comparison} compares classical, neural, and hybrid catheter models, including GRUs and backlash--LSTM combinations; and~\cite{cho2024} predicts robot shape from current and prior tendon configurations.
Residual formulations reduce what must be learned by retaining a nominal model. \cite{jiang2026residual} learns tendon-space residuals over a finite-element digital twin in an inverse-control setting.
Such models are typically trained for a single configuration, so a change in shaft routing is accommodated by retraining rather than by adaptation.

We address both effects in a single estimator driven by motor measurements alone.
Instead of reconstructing the shaft geometry or assuming a fixed functional form for its effect, an initial multidirectional sweep is encoded into a shaft-configuration context vector that conditions a recurrent residual.
The same sweep calibrates the geometric model the residual corrects, so a new configuration requires only a repeated sweep rather than retraining.

\section{Method}\label{sec:method}
In this section, we first formulate the tip-estimation problem.
We then introduce a multidirectional sweep that supports shaft-configuration context encoding and geometric catheter-model calibration.
Finally, the motor-history-conditioned GRU combines the fixed sweep context with evolving actuation history to estimate tip position.

\subsection{Problem formulation}\label{sec:problem}
We consider a tendon-driven catheter with four degrees of freedom (DOFs): two tip-bending DOFs, shaft rotation $\theta_t$, and linear insertion $\ell_t$.
The measured motor angles $\mathbf{q}_t=(\phi_{1,t},\phi_{2,t})^\top\in\mathbb{R}^2$ drive anterior--posterior and left--right tip bending.
Let $W$ be the task-space reference frame and $B$ a frame attached near the proximal end of the bending section, which moves with shaft rotation and insertion.
We estimate the tip position expressed in $B$,
\begin{equation}
 \mathbf{p}_t\equiv{}^B\mathbf{p}_{T,t}
 =({}^W R_{B,t})^\top
   ({}^W\mathbf{p}_{T,t}-{}^W\mathbf{p}_{B,t}),
 \label{eq:frame}
\end{equation}
so that $\theta_t$ and $\ell_t$ displace $B$ together with the bending section, and we focus on the response of the tip to $\mathbf{q}_t$.
Let $\xi$ denote the shaft configuration, $\mathcal{S}_\xi$ its initial multidirectional sweep, and $\mathcal{H}_t$ the processed motor-angle and drive-torque history.
For a fixed $\xi$, the estimator combines a geometric backbone $g_\xi$ and a motor-history-conditioned residual $r_\xi$:
\begin{equation}
 \hat{\mathbf{p}}_t=g_\xi(\mathbf{q}_t)+r_\xi(\mathcal{H}_t).
 \label{eq:estimator}
\end{equation}
The following subsections describe how the initial sweep $\mathcal{S}_\xi$ calibrates the geometric backbone $g_\xi$ and how the motor-history-conditioned GRU realizes the residual $r_\xi$.

\subsection{Geometric backbone and init sweep strategy}\label{sec:backbone}
The backbone $g_\xi$ maps motor angles to a nominal tip position, and we consider two geometric models for it: a single-section piecewise-constant-curvature (PCC) model, and a per-direction linear model (PDL) that relaxes its shared-arc assumption.
Both are calibrated for the current shaft configuration from an initial multidirectional sweep.

\textbf{PCC backbone.}
The single-section piecewise-constant-curvature model is
\begin{equation}
 \mathbf{u}=C\frac{\mathbf{q}-\mathbf{q}_0}{S_m}+\mathbf{u}_0,
 \qquad \alpha=\lVert\mathbf{u}\rVert_2,
 \label{eq:pcc-input}
\end{equation}
\begin{equation}
 g_{\mathrm{PCC},\xi}(\mathbf{q})=
 L R_g
 \begin{bmatrix}
 u_1(1-\cos\alpha)/\alpha^2\\
 u_2(1-\cos\alpha)/\alpha^2\\
 \sin\alpha/\alpha
 \end{bmatrix}.
 \label{eq:pcc}
\end{equation}
Here $\mathbf{u}$ contains the bending coordinates, $L$ is the effective bending length, and $R_g$ is a fitted rotation.
$C$, $\mathbf{q}_0$, $\mathbf{u}_0$, and $S_m$ denote motor coupling, motor reference, resting-curvature offset, and motor-angle scale, respectively.
The fractions use their continuous limits at $\alpha=0$.

\textbf{PDL backbone.}
Under a bent shaft, direction-dependent tendon routing can produce responses that a shared arc and a single affine coupling map do not adequately represent.
The per-direction linear model (PDL) instead fits tip inclination, displacement, and azimuth along each bending ray, then interpolates between rays:
\begin{align}
 \gamma(\rho,\psi)&=a_\gamma(\psi)\rho+b_\gamma(\psi)\rho^2,\nonumber\\
 d(\rho,\psi)&=a_d(\psi)\rho+b_d(\psi)\rho^2,\nonumber\\
 \varphi(\rho,\psi)&=\varphi_0(\psi)+s_\varphi(\psi)(\rho-\rho_0).
 \label{eq:pdl-parameters}
\end{align}
Here $\rho=\lVert\mathbf{q}\rVert_2$ and $\psi=\operatorname{atan2}(\phi_2,\phi_1)$ are motor-input magnitude and direction; $\gamma$ is twice the tip-vector inclination, $d$ the displacement from the neutral tip, and $\varphi$ the azimuth. Linearity refers to the per-ray response functions being linear in their coefficients; their dependence on motor-input magnitude and the resulting three-dimensional reconstruction are nonlinear.
The fitted coefficients $a_\cdot,b_\cdot$ govern inclination and displacement, while $\varphi_0$ and $s_\varphi$ specify azimuth at $\rho_0$ and its slope.
The $\rho^2$ terms allow the response slope to vary with motor-input magnitude while retaining a compact per-direction model.
Coefficients are interpolated linearly between neighboring rays, with periodic continuation and azimuth unwrapping.
The resulting tip direction and neutral-tip displacement reconstruct $g_\xi(\mathbf{q})$.

\textbf{Initial multidirectional sweep.}
To identify the response associated with an unknown shaft configuration, we develop an initial multidirectional sweep strategy that serves two purposes.
With $\xi$ held fixed, outward-and-return bending sweeps $\mathcal{S}_\xi$ provide \textbf{a)} paired motor angles and base-relative reference tip positions, derived from an electromagnetic (EM) tracking system, to calibrate the geometric backbone $g_\xi$, and \textbf{b)} motor-angle and drive-torque history for the shaft-configuration context $\mathbf{z}_\xi$.
The geometric parameters $\boldsymbol{\eta}_\xi$ are fitted to the paired sweep samples by damped Gauss--Newton least squares.
The EM tracking system provides reference measurements for geometric calibration, supervised training, and evaluation; it is not an input to the context encoder or the subsequent motion estimator.
These parameters and $\mathbf{z}_\xi$ remain fixed during subsequent motion with the same shaft configuration.

\subsection{Motor-history-conditioned GRU residual}\label{sec:gru}

\begin{figure}[!t]\centering
\includegraphics[width=8cm]{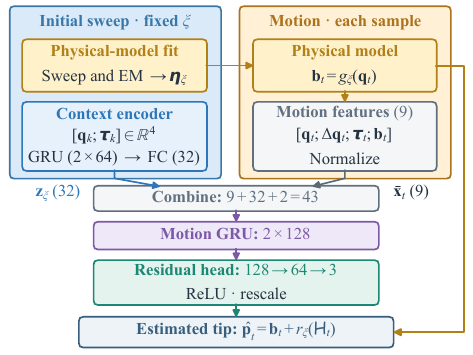}
\caption{Motor-history-conditioned GRU residual estimator.
The initial sweep $\mathcal{S}_\xi$ supplies motor history for context $\mathbf{z}_\xi$ and paired reference EM tip positions for physical-model parameters $\boldsymbol{\eta}_\xi$.
During motion, $g_\xi$ produces the nominal prediction from motor angles, and the GRU adds a history-dependent residual.}
\label{fig:gru}
\end{figure}

The GRU predicts the residual $r_\xi(\mathcal{H}_t)$ from the fixed context $\mathbf{z}_\xi$ and the evolving motor history (Fig.~\ref{fig:gru}).
Its supervised target is $\mathbf{e}_{\xi,t}=\mathbf{p}_t-\mathbf{b}_{\xi,t}$, the deviation of the measured tip position from the nominal prediction $\mathbf{b}_{\xi,t}=g_\xi(\mathbf{q}_t)$.

Before motion begins, the sweep is encoded once.
Sweep features $\mathbf{s}_{\xi,k}=[\mathbf{q}_{\xi,k}^{\top},\boldsymbol{\tau}_{\xi,k}^{\top}]^{\top}$ collect motor angles and drive-reported torques, and the encoder produces
\begin{equation}
 \mathbf{z}_\xi=\operatorname{Enc}_{\boldsymbol{\omega}}(\bar{\mathbf{s}}_\xi),
 \label{eq:context}
\end{equation}
where $\bar{\mathbf{s}}_\xi$ denotes the normalized sweep sequence subsampled at a fixed stride.
The context encoder $\operatorname{Enc}_{\boldsymbol{\omega}}$ summarizes this sequence into the fixed-length vector $\mathbf{z}_\xi$.

During motion, the nominal prediction and motor-derived channels form the per-step feature
\begin{equation}
 \mathbf{x}_{\xi,t}=
 [\mathbf{q}_t^{\top},\,
  (\Delta\mathbf{q}_t)^{\top},\,
  \boldsymbol{\tau}_t^{\top},\,
  \mathbf{b}_{\xi,t}^{\top}]^{\top}\in\mathbb{R}^{9},
 \label{eq:features}
\end{equation}
so that $\mathcal{H}_t$ collects the motor angles $\mathbf{q}_t$, per-sample angular differences $\Delta\mathbf{q}_t$, and torques $ \boldsymbol{\tau}_t$ supplied to successive recurrent updates.
Both $\mathbf{x}$ and $\mathbf{s}$ are normalized featurewise by fixed training means and standard deviations.

At each motion sample, the recurrent core receives the normalized feature and the repeated context:
\begin{align}
 \mathbf{v}_{\xi,t}&=[\bar{\mathbf{x}}_{\xi,t};\mathbf{z}_\xi],\nonumber\\
 \mathbf{h}_{\xi,t}&=\operatorname{GRU}_{\boldsymbol{\omega}}
   (\mathbf{v}_{\xi,t},\mathbf{h}_{\xi,t-1}),
 \label{eq:core}\\
 \hat{\mathbf{e}}_{\xi,t}&=\boldsymbol{\sigma}_e\odot
   H_{\boldsymbol{\omega}}(\mathbf{h}_{\xi,t}),
 \label{eq:head}
\end{align}
where $H_{\boldsymbol{\omega}}$ is a nonlinear output head and $\boldsymbol{\sigma}_e$ rescales its output to unify the units.
The hidden state is initialized at the start of each motion sequence.

The context vector and geometric parameters remain fixed for the shaft configuration, while $\mathbf{h}_{\xi,t}$ evolves at each sample.
The geometric prediction therefore enters both the feature vector and the additive estimator in \eqref{eq:estimator}.
Training settings are given in Section~\ref{sec:implementation}.

\section{Experiments and Results}\label{sec:experiments}

\subsection{Implementation Details}\label{sec:implementation}
\begin{figure}[!t]\centering
\begin{minipage}{8cm}\centering
\includegraphics[width=\linewidth]{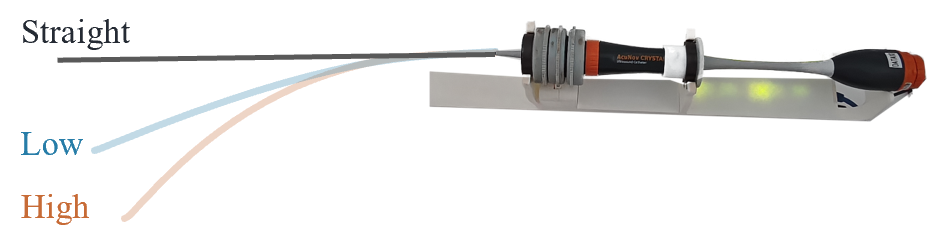}\par
{\fontsize{8}{9}\selectfont\textbf{(a) Top view}\par}
\end{minipage}\par\smallskip
\begin{minipage}{8cm}\centering
\includegraphics[width=\linewidth]{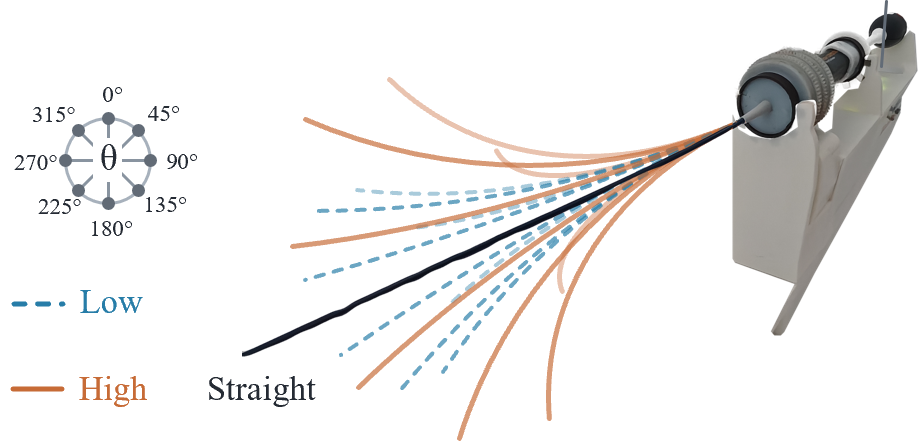}\par
{\fontsize{8}{9}\selectfont\textbf{(b) Side view}\par}
\end{minipage}
\caption{Catheter shaft configurations with two bending levels.
(a) Top view with a straight reference and low- and high-bending shaft configurations.
(b) Side view with eight bending directions at two qualitative bending levels.
The inset indexes shaft installation rotation.}
\label{fig:system}
\end{figure}

Fig.~\ref{fig:system} illustrates the 4-DOF catheter platform and shaft-bending configurations.
The top view in Fig.~\ref{fig:system}(a) shows the low- and high-bending shaft configurations.
The side view in Fig.~\ref{fig:system}(b) illustrates the nominal configuration design: eight evenly spaced shaft-bending directions at two qualitative bending levels and one straight reference.
The bending magnitude is held fixed during each recording and visually matched across catheters.
The two tip-bending motors provide measured angles and drive-reported torque signals.
Two six-DOF EM tracking sensors are attached at the distal tip and near the proximal end of the bending section to provide reference pose measurements. These measurements provide reference labels for geometric calibration, supervised training, and evaluation; subsequent tip-position estimation
uses motor-side measurements alone.

\begin{figure}[!t]\centering
\begin{minipage}[t][4cm][t]{4cm}\centering
\vspace{0pt}
\includegraphics[width=\linewidth]{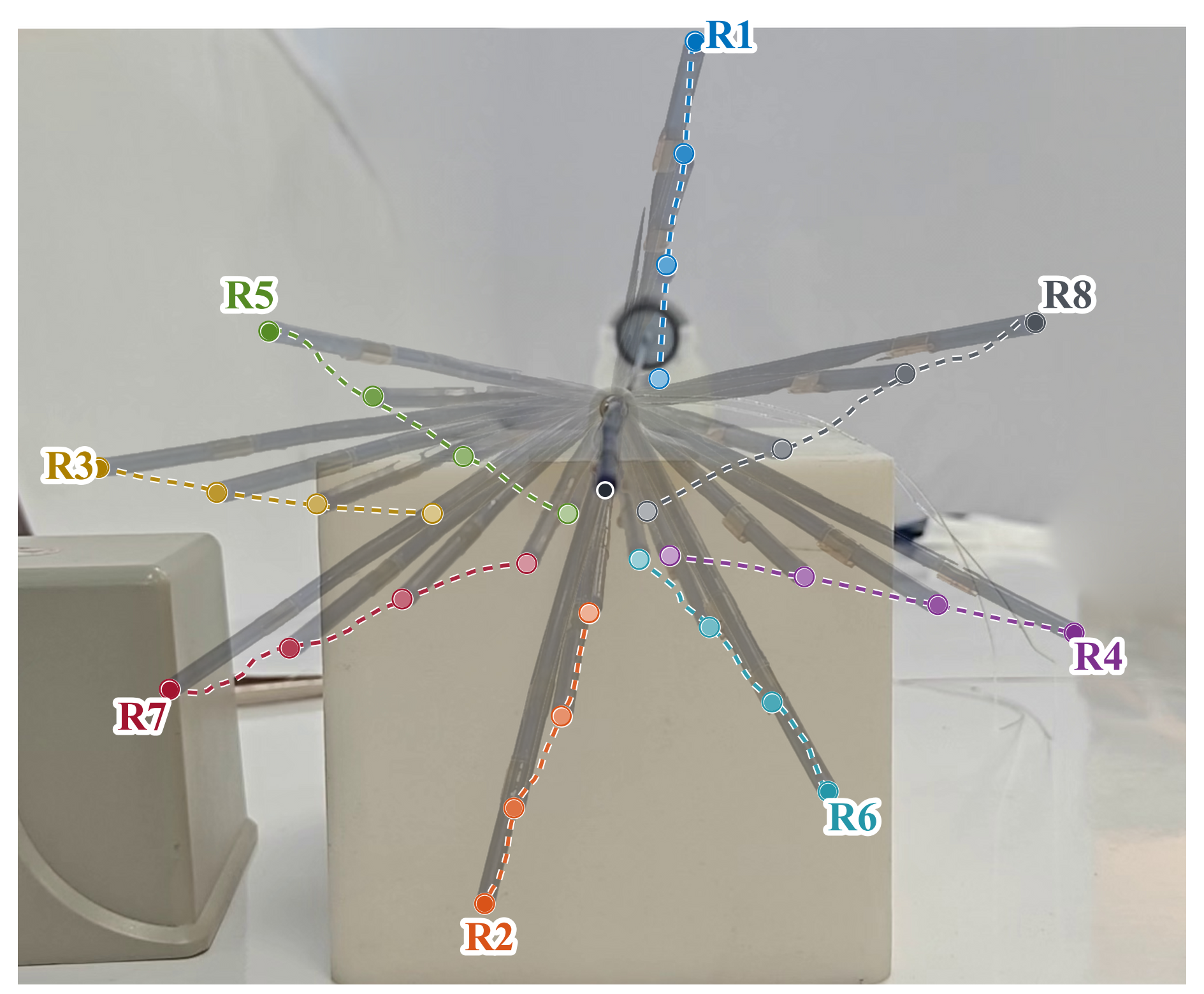}
\end{minipage}%
\begin{minipage}[t][4cm][t]{4cm}\centering
\vspace{0pt}
\includegraphics[width=\linewidth]{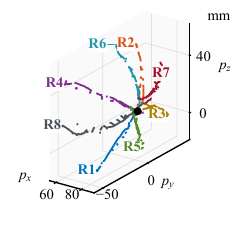}
\end{minipage}\par
{\fontsize{7}{8}\selectfont
\makebox[4cm]{\textbf{(a) Photographic trajectory}}%
\makebox[4cm]{\textbf{(b) EM sensor trajectory}}\par}
\caption{Illustration of eight-ray initial sweep.
The sweep starts from the neutral tip position (center).}
\label{fig:sweep}
\end{figure}

Data were collected from four disposable catheters of the same family, denoted C1--C4.
Each catheter dataset contains 51 runs across 17 shaft configurations, with three repeated runs per configuration.
The straight configuration serves as a reference; all reported results use the 16 bent configurations, i.e., 48 runs per catheter.
Each run begins with an eight-ray bending sweep (Fig.~\ref{fig:sweep}) followed by a 100-step random walk of both bending motors, typically yielding 6000--6300 samples recorded at 10\,Hz.

Within each catheter, all repeats of one shaft setting are held out together, and the GRU is fine-tuned on the remaining settings.
The held-out recordings supply only their initial sweeps for calibration, while the designated motion segment is reserved for testing.
The PDL backbone and GRU architecture are retained across training and prediction.
For each outer evaluation fold, the test recordings are excluded from pretraining, fine-tuning, normalization fitting, and hyperparameter selection.
In leave-one-catheter-out evaluation, these exclusions apply to every recording from the held-out catheter.
Adam uses initial learning rates of $10^{-3}$ for pretraining and $5\times10^{-4}$ for fine-tuning to minimize the masked normalized-residual loss
\begin{equation}
 \mathcal{L}(\boldsymbol{\omega})=
 \frac{\sum_{j,t}m_{j,t}
 \left\lVert F_{\boldsymbol{\omega},j,t}
   -\mathbf{e}_{j,t}\oslash\boldsymbol{\sigma}_e\right\rVert_2^2}
 {3\sum_{j,t}m_{j,t}},
 \label{eq:loss}
\end{equation}
where $j$ indexes recordings. The binary mask $m_{j,t}$ equals 1 for retained sample $t$ in recording $j$ and 0 for padded positions, excluding padding from the training loss. $F_{\boldsymbol{\omega},j,t}$ is the normalized network output for target residual $\mathbf{e}_{j,t}=\mathbf{p}_{j,t}-g_{\xi_j}(\mathbf{q}_{j,t})$; $\xi_j$ and $\mathcal{S}_j$ denote the recording's shaft configuration and initial sweep.
The applied residual is limited to $\pm25$\,mm per coordinate.
Tip accuracy is reported as mean per-run 3-D position RMSE with equal run weights, averaging learned-model errors over three seeds.
Position-error statistics are reported to one decimal place and percentages to whole numbers, with all comparisons calculated from unrounded values.

\subsection{Catheter-tip estimation results}\label{sec:tip-results}
\subsubsection{Tip estimation accuracy}
\begin{figure}[!t]\centering
\setlength{\abovecaptionskip}{3pt}
\includegraphics[width=8cm]{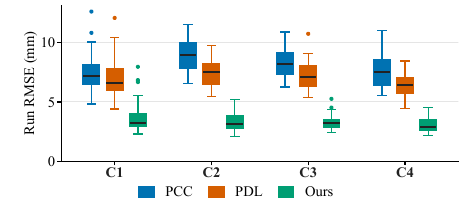}\par\nointerlineskip
\caption{Tip RMSE for PCC, PDL, and Ours (PDL+GRU).
Boxes show medians and quartiles with outliers retained.}
\label{fig:main-boxes}
\end{figure}

Across all recordings, our method achieves 3.3\,mm pooled mean per-run RMSE, compared with 8.1\,mm for PCC and 7.0\,mm for PDL (Table~\ref{tab:main} and Fig.~\ref{fig:main-boxes}).
Here PCC and PDL denote their respective full initial-sweep refits, and \emph{Ours} denotes the proposed PDL+GRU estimator with explicit motor-history residual features.
Our method reduces RMSE by 4.7\,mm (59\%) relative to PCC and 3.7\,mm (53\%) relative to PDL, improving every recording over PCC and nearly every recording over PDL.
The PDL--PCC gap (1.1\,mm) reflects the benefit of the PDL backbone, whereas adding the residual pipeline reduces RMSE by a further 3.7\,mm on the same PDL backbone.

\begin{table}[t]
\centering
\caption{Tip-position RMSE (mm).}
\label{tab:main}
\small
\setlength{\tabcolsep}{3pt}
\begin{tabular}{@{}lccccc@{}}
\toprule
Method & C1 & C2 & C3 & C4 & Pooled \\
\midrule
PCC             & 7.4 & 9.0 & 8.3 & 7.7 & 8.1 \\
PDL             & 6.9 & 7.4 & 7.2 & 6.5 & 7.0 \\
PDL+GRU            & \textbf{3.6} & \textbf{3.3} & \textbf{3.3} & \textbf{3.1} & \textbf{3.3} \\
\bottomrule
\end{tabular}
\end{table}

\subsubsection{Error distribution and motor-input magnitude}
\begin{figure}[t]\centering
\setlength{\abovecaptionskip}{3pt}
\includegraphics[width=8cm]{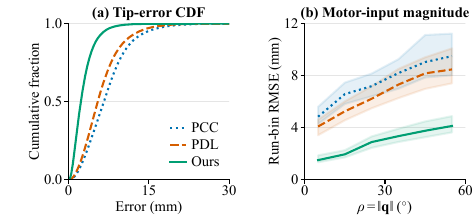}\par\nointerlineskip
\caption{Comparison of error distribution of PCC, PDL, and Ours (PDL+GRU).
(a) Tip-error cumulative distribution.
(b) Tip-error distribution at different motor-input magnitudes.}
\label{fig:est}
\end{figure}

Fig.~\ref{fig:est}(a) shows the cumulative distribution function (CDF) of per-sample tip error, pooled with equal weight per run.
At 5\,mm error, the CDF reaches 88\% for PDL+GRU, i.e., 88\% of samples have error below 5\,mm, compared with 34\% for PCC and 44\% for PDL.
These percentages average the within-run fractions, giving each recording equal weight.
This distributional shift indicates that the improvement is not limited to the largest errors.
Fig.~\ref{fig:est}(b) shows how tip-position RMSE varies with motor-input magnitude.
For PDL+GRU, median run-bin RMSE increases from 1.5\,mm in the $0$--$10^\circ$ motor-input magnitude bin to 4.1\,mm in $50$--$60^\circ$, compared with 9.5\,mm for PCC and 8.5\,mm for PDL in the highest bin (Fig.~\ref{fig:est}(b)).
The lower errors across the displayed range support the residual correction, while the remaining magnitude dependence indicates that large motor excursions are still more difficult to estimate.

\subsubsection{Random motion example}
\begin{figure}[!t]\centering
\setlength{\abovecaptionskip}{3pt}
\includegraphics[width=8cm]{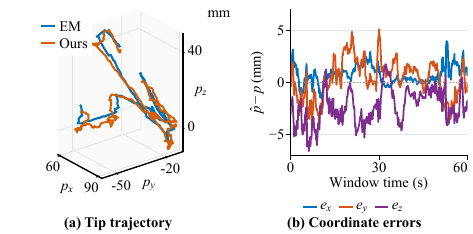}\par\nointerlineskip
\caption{Random motion example.
(a) EM (blue) and Ours (PDL+GRU, orange) on equal metric scales.
(b) Signed coordinate errors, $\hat{p}-p$.}
\label{fig:motion-example}
\end{figure}

Fig.~\ref{fig:motion-example} shows a $60$\,s segment of a C4 run, selected for spatial coverage independently of prediction error. PDL+GRU achieves 3.6\,mm RMSE over all recorded samples, spanning 31.0, 41.4, and 52.1\,mm in the measured $x$, $y$, and $z$ coordinates.
The estimate follows a spatially extended trajectory, while the signed coordinate errors show the remaining local deviations.

\subsubsection{Error over motion time}
\begin{figure}[!htbp]\centering
\setlength{\abovecaptionskip}{3pt}
\includegraphics[width=8cm]{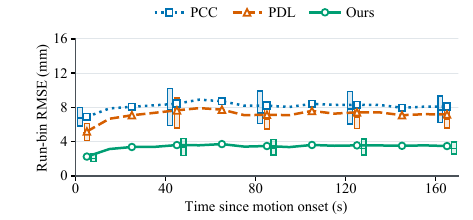}\par\nointerlineskip
\caption{Time-resolved RMSE for PCC, PDL, and Ours (PDL+GRU).
Curves show mean $10$\,s run-bin RMSE; sparse boxes show between-run quartiles and medians.
}
\label{fig:time-error}
\end{figure}

The estimation error over time is displayed in Fig.~\ref{fig:time-error}.
Over $0$--$170$\,s of motion, PDL+GRU has mean run-bin RMSE of 2.3\,mm in the first $10$\,s bin and 3.5\,mm in the last, compared with 6.9--8.1\,mm for PCC and 5.2--7.2\,mm for PDL.
Our method has a smaller between-run inter-quartile range than either geometric baseline at every displayed box group.
These results indicate lower typical error and less between-run dispersion over the displayed motion, and no pronounced drift over the displayed motion.

\subsection{Ablation studies}\label{sec:ablations}
The ablations are conducted to investigate the influence of residual-estimator choice, how much torque and sweep conditioning contribute to the model, and the ability of our method to transfer to a new catheter.

\subsubsection{Residual-estimator comparison}
\begin{figure}[t]\centering
\setlength{\abovecaptionskip}{3pt}
\includegraphics[width=8cm]{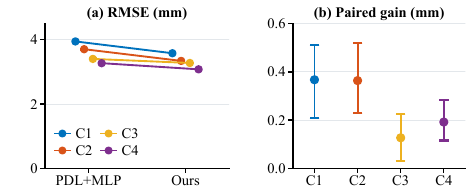}\par\nointerlineskip
\caption{RMSE comparison between different model structures: PDL+MLP and Ours (PDL+GRU).
(a) Mean run RMSE.
(b) Paired error reductions.}
\label{fig:memory}
\end{figure}

We compare two model structures to determine whether GRU improves accuracy over an MLP given the same backbone and history/calibration inputs.
PDL+MLP denotes a multilayer-perceptron residual using the same PDL backbone and motor-history and sweep context as PDL+GRU.
PDL+MLP achieves 3.6\,mm pooled RMSE, compared with 3.3\,mm for PDL+GRU, a paired reduction of 0.3\,mm (7\%; 95\% interval 0.2--0.3\,mm; Fig.~\ref{fig:memory}).
PDL+GRU lowers the mean error for each catheter and improves 81\% of recordings.
The GRU is favored, but the gain is small compared with the gain over the geometric model alone. Because both residual learners receive the same explicit history features ($\Delta\mathbf{q}_t$ and the sweep context), this comparison concerns the learner architecture rather than the presence of temporal memory.

\subsubsection{Motor torque and initial-sweep context}
\begin{figure}[t]\centering
\setlength{\abovecaptionskip}{3pt}
\includegraphics[width=8cm]{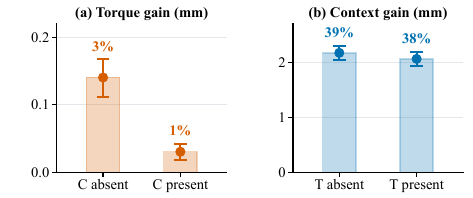}\par\nointerlineskip
\caption{Conditional RMSE reductions from (a) torque (T) and (b) sweep context (C), with setting-cluster 95\% intervals.}
\label{fig:conditioning}
\end{figure}

We remove torque and sweep context to separate the contributions of motor-load information and configuration-specific calibration.
Pooled RMSE is 5.5\,mm without torque or   context, 5.4\,mm with torque alone, 3.4\,mm with   context alone, and 3.3\,mm with both (Table~\ref{tab:conditioning}).
With context present, torque reduces error by less than 0.1\,mm (1\%); without context, its reduction is 0.1\,mm (3\%; Fig.~\ref{fig:conditioning}(a)).
Adding context reduces error by 2.1\,mm  with torque and 2.2\,mm without torque (Fig.~\ref{fig:conditioning}(b)).
The pooled paired intervals are all positive, but the torque benefit is much smaller than the benefit of sweep conditioning in this training setup.

\begin{table}[t]\centering
\caption{Torque/context ablation RMSE (mm).}
\label{tab:conditioning}
\footnotesize
\setlength{\tabcolsep}{2.5pt}
\begin{tabular}{@{}ccrrrrr@{}}
\toprule
Torque & Context & C1 & C2 & C3 & C4 & Pooled \\
\midrule
No & No & 5.4 & 5.8 & 5.5 & 5.4 & 5.5 \\
Yes & No & 5.3 & 5.6 & 5.4 & 5.2 & 5.4 \\
No & Yes & 3.6 & 3.4 & 3.3 & 3.1 & 3.4 \\
Yes & Yes & \textbf{3.6} & \textbf{3.3} & \textbf{3.3} & \textbf{3.1} & \textbf{3.3} \\
\bottomrule
\end{tabular}
\end{table}

\subsubsection{Model adaptation to new catheters}
\begin{figure}[t]\centering
\setlength{\abovecaptionskip}{3pt}
\includegraphics[width=8cm]{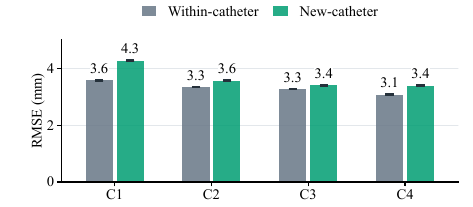}\par\nointerlineskip
\caption{PDL+GRU under within-catheter and held-out-catheter evaluation.
Held-out sweeps supply calibration/context only; learned-model training excludes the test catheter.}
\label{fig:new-catheter}
\end{figure}

We withhold each catheter in turn to test whether initial-sweep calibration can support transfer without training the residual learner on the new catheters.
For each held-out catheter, the same feature design and GRU model family are trained only on the other catheters.
The held-out catheter's initial sweeps are used only to refit $g_\xi$ and to compute $\mathbf{z}_\xi$; they never train or fine-tune the GRU.
Leave-one-catheter-out testing gives 3.7\,mm pooled RMSE, compared with 3.3\,mm under within-catheter setting-held-out evaluation (Fig.~\ref{fig:new-catheter}).
The held-out C1--C4 errors are 4.3, 3.6, 3.4, and 3.4\,mm, respectively, with 74\% of recordings below 4\,mm after averaging seed-specific errors.
These results support transfer to unseen catheters of the same family with a modest pooled penalty.

\subsection{Discussion}\label{sec:discussion}
The reduction from 7.0\,mm with PDL to 3.3\,mm with PDL+GRU shows that the context- and history-conditioned residual adds substantial predictive value beyond the geometric model alone.
The smaller 0.3\,mm advantage over PDL+MLP suggests that residual-learner choice is a secondary factor within the shared feature representation.
These comparisons support combining a calibrated nominal model with a learned correction, but do not isolate recurrent memory or identify a unique physical cause of the residual error.
The 2.1\,mm gain from context with torque, compared with a torque gain below 0.1\,mm when context is available, makes configuration-specific calibration the more influential conditioning source in this setup.
Reference tip measurements remain necessary during initial calibration, although subsequent prediction uses motor measurements only.
New-catheter RMSE increases from 3.3 to 3.7\,mm, indicating useful transfer without GRU fitting on the held-out catheter.

Several factors bound the scope of these results.
The estimator requires reference tip positions during the initial sweep, shifting the sensing requirement from continuous operation to initial calibration rather than removing it.
The shaft configuration is assumed fixed after the sweep; a repositioned shaft invalidates the context and geometric parameters and requires a new sweep, and detecting such changes online is not addressed.
Transfer was only evaluated retrospectively on four units of one device family.
Larger motor-input magnitudes are associated with higher run-bin RMSE (Fig.~\ref{fig:est}(b)), indicating reduced accuracy for larger actuator excursions.

\section{Conclusion}\label{sec:conclusion}
This paper presents a residual tip-position estimator that combines a geometry-based backbone with a GRU residual conditioned on an initial-sweep context encoding the shaft configuration and on the motor history capturing hysteresis.
Across the evaluated bent-shaft recordings, it achieves a mean per-run 3-D tip-position RMSE of 3.3\,mm.
This represents an RMSE reduction of 53\% relative to PDL and 59\% relative to PCC.
The ablations show that the shaft-configuration context derived from the initial sweep is the dominant contributor to accuracy, while replacing the MLP residual with a GRU or adding motor torque yields smaller gains.
Leave-one-catheter-out evaluation gives 3.7\,mm pooled RMSE, indicating the model's ability to adapt to new catheters.
Future work should address reduced calibration budgets and prospective evaluation on additional catheters.

\section*{Acknowledgments}
In this paper, ChatGPT was used to help implement the described algorithms, analysis code, and plotting scripts. 
Claude was used to polish portions of the manuscript text, including restructuring and rewording for clarity and continuity.

\section*{Disclaimers}
The concepts and information presented in this paper are based on research results that are not commercially available. Future availability cannot be guaranteed.

\bibliographystyle{IEEEtran}
\bibliography{citation}

\end{document}